\documentclass{IEEE_lsens}

\usepackage{textcomp}

\usepackage[noadjust]{cite}

\ifCLASSINFOpdf

\else

\fi

\usepackage[T1]{fontenc} 
\usepackage{amsmath}

\usepackage{algorithm} 
\usepackage[algo2e]{algorithm2e} 
\usepackage{graphicx}
\usepackage{textcomp}
\usepackage{url}
\usepackage{hyperref}
\usepackage[cmintegrals]{newtxmath}
\usepackage{bm}
\usepackage{array}
\usepackage{url}
\usepackage{cite}

\ifCLASSINFOpdf
\else
\fi
\providecommand{\hypersetup}[1]{\relax}

\hypersetup{pdftitle={Submitted},
pdfsubject={Typesetting},
pdfauthor={Michael D. Shell},
pdfkeywords={Class, IEEE, IEEE\_lsens, IEEE Sensors Letters, LaTeX, Typesetting, TeX}}

\def\BibTeX{{\rm B\kern-.05em{\sc i\kern-.025em b}\kern-.08em
    T\kern-.1667em\lower.7ex\hbox{E}\kern-.125emX}}
    
\begin{document}


\IEEELSENSarticlesubject{The Pre-print, Submitted to ICCAS}

\title{Epersist: A Self Balancing Robot Using PID Controller And Deep Reinforcement
Learning}

%
\author{\IEEEauthorblockN{Ghanta Sai Krishna\IEEEauthorrefmark{1}\IEEEauthorieeemembermark{1}, Garika Akshay\IEEEauthorrefmark{1},
Dyavat Sumith\IEEEauthorrefmark{1}}
\IEEEauthorblockA{\IEEEauthorrefmark{1}Department of Data Science and Artificial Intelligence,
IIIT Naya Raipur, Chattisgarh, 492101, India\\
\IEEEauthorieeemembermark{1}Student Member, IEEE}

\thanks{Corresponding author: Ghanta Sai Krishna  (e-mail: ghanta20102@iiitnr.edu.in)\protect\\}
\thanks{Copy righted Preprint}%
\thanks{Associate Editor: }%
\thanks{Digital Object Identifier}}

\IEEEtitleabstractindextext{%
\begin{abstract}
A two-wheeled self-balancing robot is an example of an inverse pendulum and is an inherently non-linear, unstable system. The fundamental concept of the proposed framework "Epersist" is to overcome the challenge of counterbalancing an initially unstable system by delivering robust control mechanisms, Proportional Integral Derivative (PID), and Reinforcement Learning (RL). Moreover, the micro-controller NodeMCU ESP32 and inertial sensor in the Epersist employ fewer computational procedures to give accurate instruction regarding the spin of wheels to the motor driver, which helps control the wheels and balance the robot. This framework also consists of the mathematical model of the PID controller and a novel self-trained advantage actor-critic algorithm as the RL agent. After several experiments, control variable calibrations are made as the benchmark values to attain the angle of static equilibrium. This "Epersist" framework proposes PID and RL-assisted functional prototypes and simulations for better utility.
\end{abstract}

}


\maketitle

\section{Introduction}
The two-wheeled self-balancing robot (TWSBR) is a standard robot with applications in various fields, including transportation and exploration. Over the last few decades, academics and industry have been paying close attention to both the design and regulation of the TWSBR. The TWSBR is considered as a high-order, multi-variable, nonlinear, tightly coupled, inherently unstable system. Conventional methods like PID\cite{b1}, fuzzy\cite{b2}, and sliding mode control\cite{b3} are proposed in the recent times. Two wheeled mobile robot balancing controller has been tackled as either linearized  or as a nonlinear model\cite{b4}. The physical characteristics of robot are crucial to achieve optimal control \cite{b5}. In practical applications It is frequently desirable to obtain optimality beyond simple stabilisation. Though the existing works had achieved maximum stabilization, they are less capable of achieving optimality. The control systems are not optimal, if the stabilization criterion's depend physical characteristics of robot. So there might be a need of advanced techniques for achieving optimality. 

In recent works, Reinforcement learning (RL) has been included in TWSBR as a control mechanism to attain optimality and achieve stability \cite{b6}. The RL enables robots to learn, adapt, and optimize their behaviours by interacting with their surroundings. RL solves the optimization issues that involve an agent interacting with its environment and changing its behaviours or control policies in response to inputs or rewards. To attain better stability, RL techniques are widely used in TWSBR, like Q-learning\cite{b7}, Proximal Policy Optimization (PPO) \cite{b8}, and Soft Actor-Critic (SAC) \cite{b9}.  However, the RL frameworks of existing solutions are based on Q-Learning, PPO and SAC. The proposed methodology highlights the benefits of the advantage Actor-Critic Algorithm (A2C) \cite{b10} based self-trained model for the self-balancing robot. 
 
Irrespective of the robot's control mechanisms, it is essential to understand the theoretical aspects like the transfer function\cite{b11}, stability\cite{b12} etc., which are crucial in judging the system's stability at a time instance. To understand these theoretical aspects of the robot, simulations of the robot are also required. However, the circumstances and the constraints in executing the hardware prototyping are slightly different from the simulations. Moreover, the hardware experimental data is valuable, and RL architectures usually require millions of data volumes \cite{b13}. Apart from the robot's functionality, it is essential to make it more cost-effective. However, the existing solutions utilize micro-controllers such as Raspberry-PI\cite{b14}, Arduino-UNO \cite{b15} etc. The overall cost of the robot by utilizing these micro-controllers is expensive. Thus there is a need to improve the cost-efficiency and optimal utility of the system. To overcome all these disadvantages in the existing solutions, the proposed methodology effectively analyses both theoretical aspects via simulations and the hardware prototyping conditions of the robot. 

. Moreover, the proposed solution utilizes Node-MCU ESP-8266 as a micro-controller to make a cost-effective system. To improve the practical utility of the robot, the proposed solution also consists of a mobile interface which connects to the micro-controller. The significant contributions of the paper are as follows :

The significant contributions and improvements from existing works of Epersist are as follows: 

\begin{description}
  \item[$\bullet$] Analyzing and deriving the theoretical aspects and effects of the PID control mechanism on the robot. 
  \item[$\bullet$] Building a novel, robust and less computational A2C algorithm-based Deep Reinforcement Learning agent.
  \item[$\bullet$] Deploying the PID mechanism and RL model in the NodeMCU to build a cost-effective hardware prototype.
 \item[$\bullet$] Deploying a Bluetooth-based mobile application to control the robot for greater utility 
\end{description}

\relax

\begin{figure*}[t]
\centerline{\includegraphics[height=50mm,width=180mm]{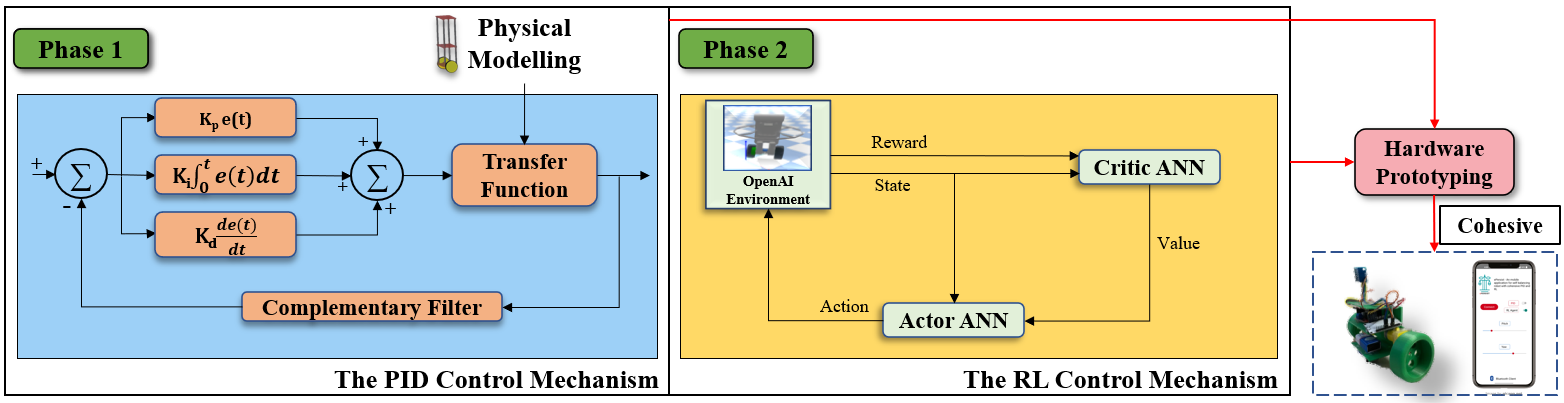}}
\caption{The Conceptual Overview of proposed framework "Epersist"}
\label{fig1}
\end{figure*}

\section{Methodology of Epersist}
This section discusses the procedures involved in the proposed methodology, which is categorized into 2 phases- Physical modelling and PID control mechanism; Deep RL agent. The overview of the proposed methodology is represented in Fig. \ref{fig1}.

\subsection{Physical Modelling and PID Control Mechanism}
\begin{figure}[h]
\centerline{\includegraphics[height=40mm,width=50mm]{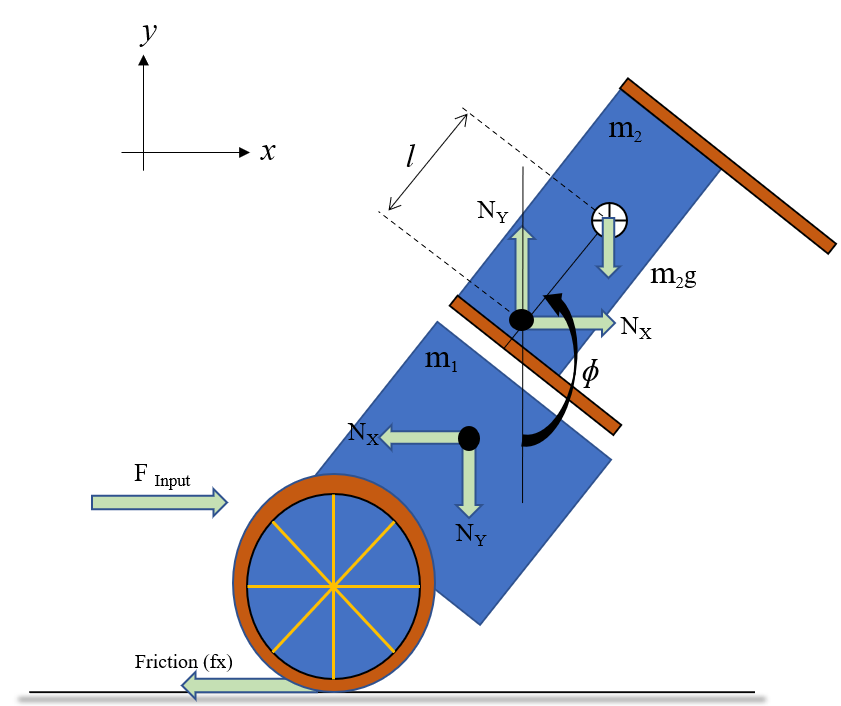}}
\caption{Free-Body Diagram of the Robot}
\label{fig2}
\end{figure}

The physical model of the robot is examined as the fundamental concept of the "Inverted Pendulum", which involves calculating the transfer function of control variables (pitch and yaw) of the robot is shown in Fig. \ref{fig2}. The robot's pitch is the angle of deviation along the Y-axis, whereas the yaw is the robot's position along the X-axis. The entire robot is divided into cart and pendulum based on the weight composition ($m_{1}$,$m_{2}$) and position, respectively. The initial conditions of mass and speed are taken to zero ($x(0) = 0$).

\begin{equation}
F_{input} =  (m_{1} + m_{2})\ddot{x} + f\dot{x} + m_{2}l\ddot{\phi}cos(\phi)-m_{2}l\dot{\phi}^{2}sin\phi \\
\label{eq1}
\end{equation}
\begin{equation}
(I_{2} + m_{2}l^{2})\ddot{\phi} + m_{2}glsin\phi = -m_{2}l\ddot{x}cos\phi
\label{eq}
\end{equation}

Where, the distance to the pendulum's centre of mass, $l$, and the angle between the pendulum and the Y-axis, $\phi$, friction is represented by the parameter $f$. The eq. \ref{eq}1, eq. \ref{eq} are derived from the fundamental law's of motion. The resultant transfer functions of both pitch and yaw for the robot are shown in eq. \eqref{eq3}, \eqref{eq4} respectively.

\begin{equation}
G_{pitch}(s) = \frac{\frac{m_{2}l}{q}s}{s^{3} + \frac{f(I_{2} + m_{2}l^{2})}{q}s^{2} - \frac{(m_{2}+m_{1})m_{1}gl}{q}s - \frac{fm_{2}gl}{q}}
\label{eq3}
\end{equation}

\begin{equation}
G_{yaw}(s) = \frac{\frac{(I_{2} + m_{2}l^{2})s^{2} - gm_{2}l}{q}}{s^{4} + \frac{f(I_{2} + m_{2}l^{2})}{q}s^{3} - \frac{(m_{2}+m_{1})m_{1}gl}{q}s^{2} - \frac{fm_{2}gl}{q}s}
\label{eq4}
\end{equation}

The transfer function helps understand the system's fundamental aspects (e.g. stability), which will be explained in further subsections. On the other hand, the 6-axis Inertial Measurement Unit (IMU) sensor calculates the 3-axis gyroscopic ($g_{x}, g_{y}, g_{z}$)and 3-axis accelerometer ($a_{x}, a_{y}, a_{z}$) measures. Computationally, the pitch and yaw calculation is based on the sensor measures, which is different from the conventional mathematical model. The instantaneous pitch angle ($\phi$) is dependent on the previous pitch angle ($\hat{\phi}(t=0) = 0$) and is calculated with a complementary filter as derived in eq. \ref{eq5}.

\begin{equation}
\phi = [\alpha(\hat{\phi} + g_{x})] + [(1-\alpha)(ATAN2(a_{y}, a_{z}))]
\label{eq5}
\end{equation}

Where $\alpha$ is the filer coefficient, and $ATAN2(a_{y}, a_{z}$ represents the angle of ($a_{y}, a_{z}$) in the plane. The error ($e(t)$) at an instance ($t$) is the difference between the current pitch angle and target pitch angle. This feedback error is applied to the PID control mechanism at every time instance as dervied in eq. \ref{eq6}. Finally, this is responsible to generate the command for instructing motor driver. For balancing the robot rigidly, the physical parameters play a crucial position. The PID calibration (tuning the $K_{p}, K_{i}, K_{d}$ values) is performed with trial-error method.  

\begin{equation}
Ouput = K_{p}e(t) + K_{i}\int e(t)dt + K_{d}\frac{\mathrm{d} e(t)}{\mathrm{d} x}
\label{eq6}
\end{equation}

\subsection{The Deep Reinforcement Learning Agent}
The novel self-trained deep reinforcement learning agent is based on the advantage Actor-Critic (A2C) algorithm, which consists of two dependent and similar neural networks (actor and critic) and integrates RL's policy and value algorithms. The actor (policy function) decides an action at each iteration, and the critic (value function) estimates the quality or the value index of a delivered initial state.

\begin{figure}[h]
\centerline{\includegraphics[height=38mm,width=75mm]{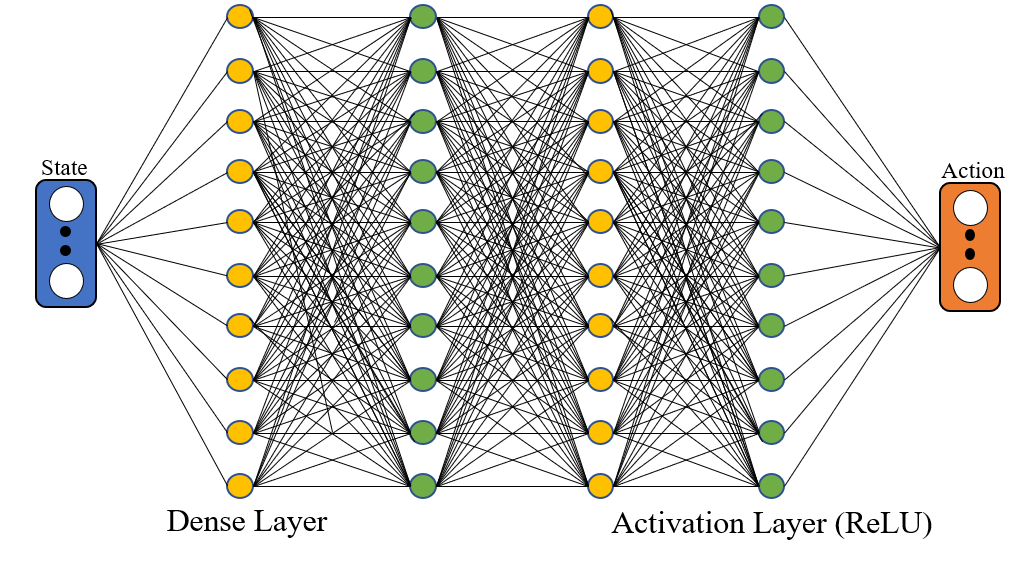}}
\caption{The Architecture of Actor ANN (Previous State - Action)}
\label{fig3}
\end{figure}

\begin{figure}[h]
\centerline{\includegraphics[height=38mm,width=75mm]{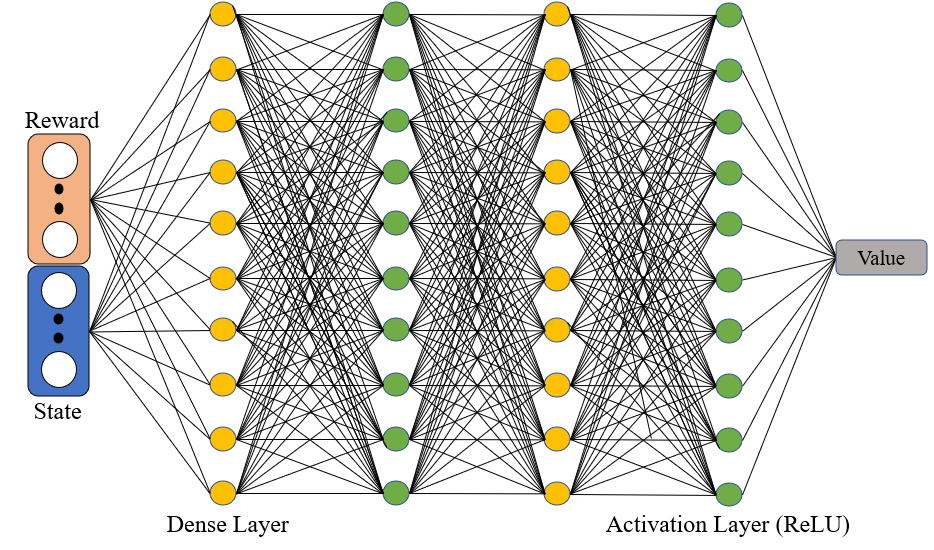}}
\caption{The Self-Trained Architecture of Critic ANN}
\label{fig4}
\end{figure}

\begin{figure}[h]
\centerline{\includegraphics[height=64.46mm,width=84mm]{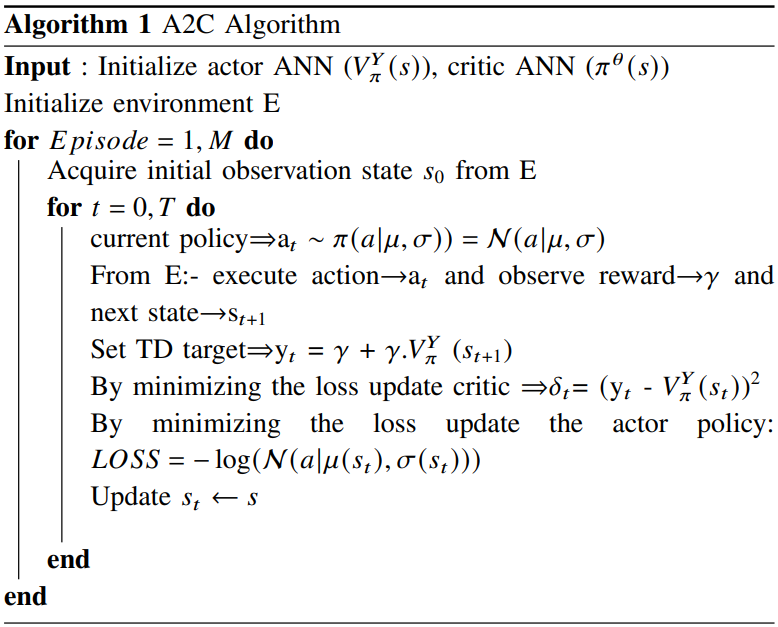}}
\label{algo}
\end{figure}

The Artificial Neural Network (ANN) architecture of both actor and critic is visualized in Fig. \ref{fig3}, \ref{fig4}. Based on the system's previous state, the actor decides the further action. Similarly, based on the previous action (rewards) and the state, the critic will produce the quality value of that action. The actor will learn from the temporal difference error (TD), calculated by the advantage function in each nested iteration. The detailed mathematical functionality of the A2C algorithm is characterised in algorithm \ref{algo}, where the initial yaw and pitch values are derived in the form of weights $Y$ and $\theta$.

\section{Experimental Results}
In this section, the experimental results are presented and analysed. Firstly, the physical modelling and PID tuning are performed based on our system design in the MATLAB Simulink. The RL model for the robot is designed in the Pybullet-OpenAI environment and further injected to the micro-controller. The physical parameters of the robot are shown in Table \ref{Table1}. 

\begin{table}[ht] 
\begin{center}
\scalebox{0.9}
{
\begin{tabular}{cr cr}
\hline
\textbf{Physical Parameter} & \textbf{Value/Unit}\\
\hline
Mass of main body & 135 g\\
Mass of pendulum & 60 g\\
Diameter of Wheel & 5 cm\\
Distance between wheels & 20 cm\\
Static Friction b/w Surfaces & 1.15\\
\hline
\end{tabular}
}
\caption{Physical parameters of Epersist Robot}
\label{Table1}
\end{center}
\end{table}

The hardware prototype proposed is cost-effective and consists of NodeMCU ESP32, IMU MPU 6050, H-1298N motor bridge, 6V gear motors, and a self-designed 3D printed base plate. The NodeMCU ESP32 acts as the micro-controller, in which the RL model or PID mechanism is manually uploaded. The injection of the RL model (.h5 extension) to NodeMCU ESP32 is one of the challenging tasks, which is handled precisely. This micro-controller is compatible with connecting with a mobile phone via Bluetooth. An interactive mobile application is built to provide the robot with initial control variables (pitch, yaw) as shown in Fig. . These interactions between the mobile application and the robot are efficient. An Inertial Measurement Unit (6-axis IMU MPU 6050) sensor is utilised to acquire the 3-axis accelerometer  and 3-axis gyroscopic   measures. These six measures are responsible for the calculation of the pitch (tilt angle) and yaw (horizontal movement) of the robot. 

\begin{figure}[h]
\centerline{\includegraphics[height=40mm,width=80mm]{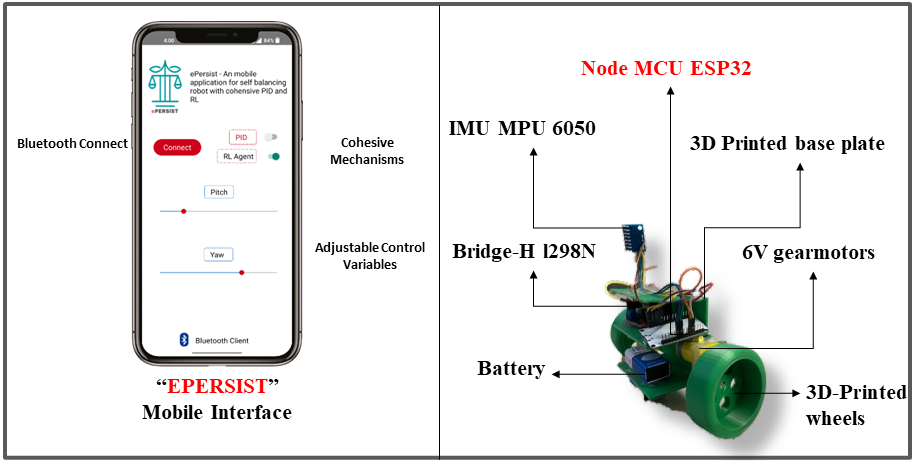}}
\caption{Epersist Robot and the Mobile Interface}
\label{fig5}
\end{figure}

The optimal calibration points must be identified to calculate the target values of control variables and sensor measures. After several experimentation's, the $K_{p}, K_{i}$ and $K_{d}$ values are tuned and are calibrated to 1970, 21950, 19.5 respectively. Furthermore, the calibrated offsets of the sensor measures are shown in Table \ref{Table2}.

\begin{table}[ht] 
\begin{center}
\scalebox{0.9}
{
\begin{tabular}{cr cr}
\hline
\textbf{Offset} & \textbf{Value}\\
\hline
X Accelerometer Offset & -1780\\
Y Accelerometer Offset & 750\\
Z Accelerometer Offset & 2700\\
X Gyroscopic Offset & 180\\
Y Gyroscopic Offset & 76\\
Z Gyroscopic Offset & 61\\
\hline
\end{tabular}
}
\caption{Offset measures for the IMU MPU 6050 Sensor}
\label{Table2}
\end{center}
\end{table}

Irrespective of the system's physical characteristics, the trained RL model can be utilized for our robot. As neural networks are involved in the RL model, the training and uploading time to NodeMCU is longer than the PID mechanism. The description and summary of the trained RL model are shown in Table \ref{Table3}.  

\begin{table}[ht] 
\begin{center}
\scalebox{0.9}
{
\begin{tabular}{cr cr}
\hline
\textbf{Parameter} & \textbf{A2C}\\
\hline
Trained Episodes & 7775\\
Trained Steps & 1.5e+06\\
Total Training Time & 19 Hrs\\
Time To Upload  & 23.5 Secs\\
Maximum Possible Reward After Training & 60\\
Maximum Achieved Reward After Training & 56.42\\
\hline
\end{tabular}
}
\caption{Summary and Description of RL self-trained RL model}
\label{Table3}
\end{center}
\end{table}

\begin{figure*}[ht]
\centerline{\includegraphics[height=40mm,width=180mm]{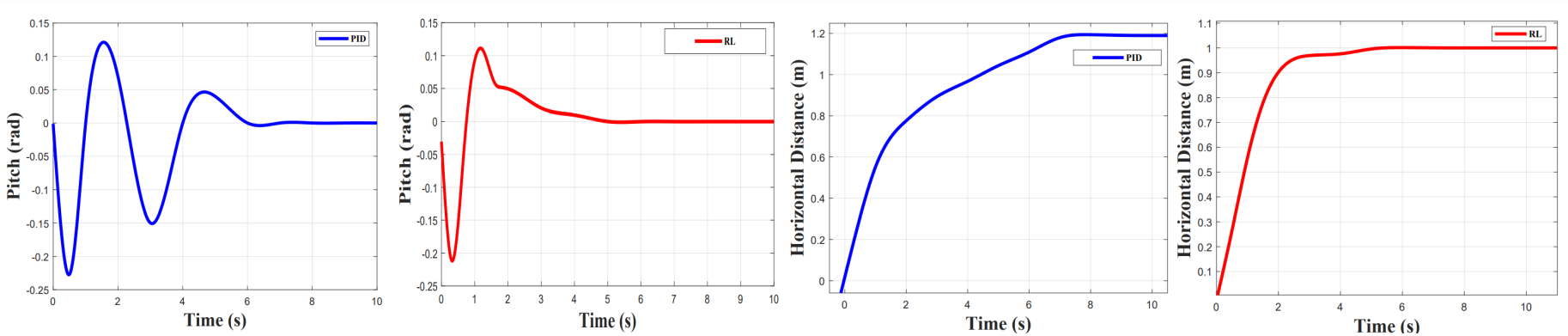}}
\caption{Comparative assessment of PID and RL mechanism - Pitch Vs Time, Yaw Vs Time}
\label{fig6}
\end{figure*}

After tuning the $K_{p}, K_{i}, K_{d}$ and hyper-parameters of both control mechanisms. The performance of the control mechanisms is analyzed with the sensor measures from the functional prototype. The comparative assessment of both control mechanisms with the hardware prototype is demonstrated in Fig. \ref{fig6}. The comparison between the PID and RL is based on the angle of deviation (pitch) and horizontal distance (yaw) covered by the robot over a while. The final angle of deviation must be approximately zero to call it a balanced system. Our experimentation's observed that the robot's overall movement with PID was not as smooth as with RL. The PID-assisted robot's angle of deviation (pitch) oscillates more than the RL. As the number of oscillations increases, the smoothness of the robot decreases in a unit of time. The mean settling time of the PID robot is 7 seconds, whereas the RL robot takes 5 seconds to achieve maximum stability. 
Furthermore, the RL robot had covered less distance when compared to the PID robot to achieve maximum stability. The adequate difference between the distance covered by the robot for both control mechanisms is approximately 20 cm. To conclude, the RL control mechanism for the robot is better and more effective than the PID control mechanism in terms of mean settling time and distance travelled in mean settling time.

\section{Conclusion}
The proposed TWSBR "Epersist" is a flawless end-to-end framework for studies of advanced control techniques due to its complicated task of balancing the structure. The proposed framework has many advantages in terms of the time complexity of the contemporary self-trained RL agent and the cost efficiency of the robot over other existing frameworks. The physical modelling of the system is appropriate, and the outcomes match the experimental results. Initially, the framework is analyzed via simulation (MATLAB for PID, OpenAI for RL), and hardware prototyping is performed with NodeMCU ESP32 micro-controller, IMU MPU 6050 sensor. The PID and RL mechanisms are analyzed from the initial stages based on the physical system design. We represented the performance and experimental results for both control mechanisms. The utility of this robot is simple with the interactive mobile interface, which is connected over Bluetooth. This robot is limited to a random path. The robot cannot move on the desired path. An extension of the desired path can be included in the framework for future work, and the advanced RL control agents can also be included.

\end{document}